\documentclass[letterpaper, 10 pt, conference]{ieeeconf}
\IEEEoverridecommandlockouts
\usepackage[hyphens]{url}
\usepackage{graphicx}
\usepackage{cite}
\usepackage[table]{xcolor}
\usepackage{caption}
\usepackage{booktabs}
\usepackage{amsmath}
\usepackage{amssymb}
\usepackage{multirow}
\usepackage{array}
\usepackage{colortbl}

\newcommand{\cmark}{\checkmark}

\definecolor{ceilingcolor}{RGB}{220,40,40}
\definecolor{floorcolor}{RGB}{45,170,20}
\definecolor{wallcolor}{RGB}{145,210,225}
\definecolor{windowcolor}{RGB}{105,155,205}
\definecolor{cabinetcolor}{RGB}{235,120,120}
\definecolor{chaircolor}{RGB}{195,195,80}
\definecolor{bedcolor}{RGB}{250,180,105}
\definecolor{sofacolor}{RGB}{135,95,180}
\definecolor{tablecolor}{RGB}{35,125,185}
\definecolor{countercolor}{RGB}{90,170,140}
\definecolor{deskcolor}{RGB}{210,145,70}
\definecolor{pillowcolor}{RGB}{230,150,205}
\definecolor{nightstandcolor}{RGB}{160,130,95}
\definecolor{toiletcolor}{RGB}{120,190,230}
\definecolor{sinkcolor}{RGB}{70,155,190}
\definecolor{bathtubcolor}{RGB}{150,195,220}
\definecolor{tvscolor}{RGB}{145,185,25}
\definecolor{boxcolor}{RGB}{190,125,70}
\definecolor{whiteboardcolor}{RGB}{180,185,190}
\definecolor{furniturecolor}{RGB}{250,120,15}
\definecolor{objectscolor}{RGB}{190,165,210}

\newcommand{\tableheaderheight}{5em}
\newcommand{\headercell}[2]{%
  \raisebox{-2em}{\makebox[#1][c]{#2}}%
}
\newcommand{\vclassheader}[2]{%
  \parbox[b][\tableheaderheight][b]{0.82cm}{%
    \centering
    \rotatebox{90}{#2}\\[-0.15ex]
    \textcolor{#1}{\rule{0.8em}{0.8em}}%
  }%
}
\newcolumntype{C}{>{\centering\arraybackslash}m{0.82cm}}
\newcolumntype{D}{>{\centering\arraybackslash}m{1.02cm}}

\title{\LARGE \bf
OVM-Occ: Open-Vocabulary Monocular Occupancy Prediction via Geometry-Aware Vision-Language Distillation
}
\author{
Shuyang Yu$^{1,2}$, Haibo Qin$^{2}$, Shaohan Hu$^{2}$, Wei Shan$^{2}$, Xiaoqing Guan$^{1}$, You Wang$^{1,\dagger}$, and Guang Li$^{1}$%
 \thanks{$^{1}$Zhejiang University. $^{2}$AgiBot.}%
\thanks{$^{\dagger}$Corresponding author: You Wang, king\_wy@zju.edu.cn.}
}

\begin{document}

\maketitle
\thispagestyle{empty}
\pagestyle{empty}

\begin{abstract}
Monocular occupancy prediction in indoor scenes aims to recover a scene's 3D geometry and semantics from a single RGB image. 
Existing methods are restricted to a closed set of predefined categories, limiting their ability to generalize to unseen categories.
This limitation is particularly critical in indoor scenes, where objects exhibit highly diverse and fine-grained semantic categories.
We study open-vocabulary monocular occupancy prediction for indoor scenes and present OVM-Occ, 
a Gaussian-based occupancy framework with an instance-aware vision-language teacher.
Specifically, we adopt Language-Embedded Gaussians as the intermediate scene representation,
propose a geometry-aware deformable attention module to refine Gaussian semantic features,
and construct instance-level vision-language embeddings to provide reliable semantic supervision.
We further introduce geometry-aware decoupled semantic distillation to mitigate the interference between geometry learning and semantic learning.
To evaluate open-vocabulary occupancy prediction performance, we introduce OpenOcc-ScanNet and OpenOcc-Replica, two extended benchmarks with more diverse and fine-grained semantic categories.
Experiments show that our method achieves state-of-the-art performance in open-vocabulary monocular occupancy prediction for indoor scenes, 
outperforming the previous state-of-the-art approach by 12.2 mIoU and 0.6 IoU on the Occ-ScanNet dataset, 
with consistent gains under new open-vocabulary evaluation protocols. Project Page: \mbox{\textcolor[HTML]{FB3199}{\url{https://genie-universal-algorithm.github.io/OVM-Occ/}}}
\end{abstract}

\section{Introduction}
A comprehensive understanding of the surrounding 3D environment is fundamental to embodied perception, robotic navigation, and physical interaction~\cite{EmbodiedIntelligence,embodiedscan}.
Occupancy prediction represents the environment as a dense volumetric grid, jointly estimating the occupancy state of each spatial element and assigning semantic labels to occupied regions~\cite{occ3d}.
This unified representation, which integrates geometry and semantics, supports fine-grained spatial reasoning and downstream embodied tasks~\cite{volumetric}.

Vision-based occupancy prediction has rapidly advanced in autonomous driving scenarios~\cite{bevformer,tpvformer,surroundocc,gaussianformer}, 
where multi-view images provide strong geometric and semantic cues.
By contrast, indoor occupancy prediction is more challenging and comparatively underexplored due to denser layouts and more severe occlusions.
These challenges become even more pronounced in the monocular setting, where 3D occupancy must be inferred from a single RGB image without multi-view constraints.
Recent methods~\cite{embodiedocc,roboocc,GPOcc} have explored Gaussian-based representations for occupancy prediction, achieving promising improvements in indoor scenes.

\begin{figure}[t]
\centering
\begin{minipage}{\columnwidth}
\centering
\includegraphics[width=1.0\columnwidth]{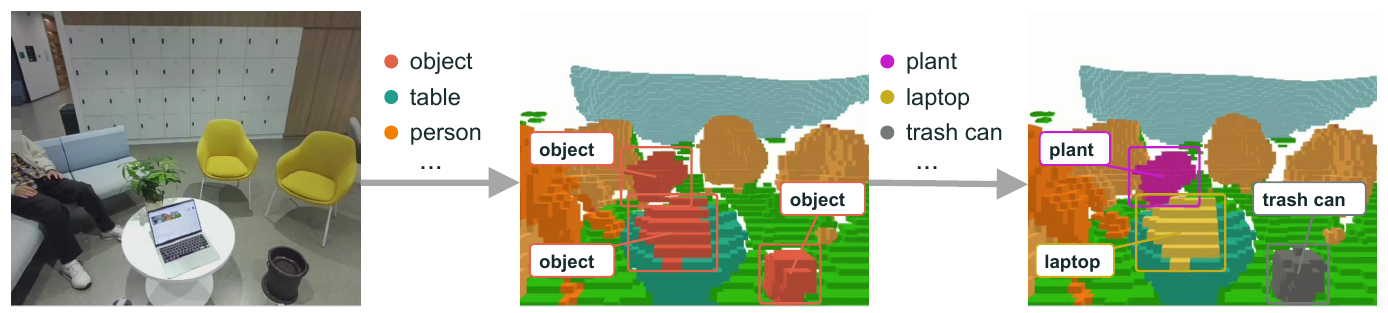}\\[-0.5ex]
\small (a) Open-Vocabulary Occupancy Prediction
\end{minipage}\\[0.8ex]
\begin{minipage}{\columnwidth}
\centering
\includegraphics[width=1.0\columnwidth]{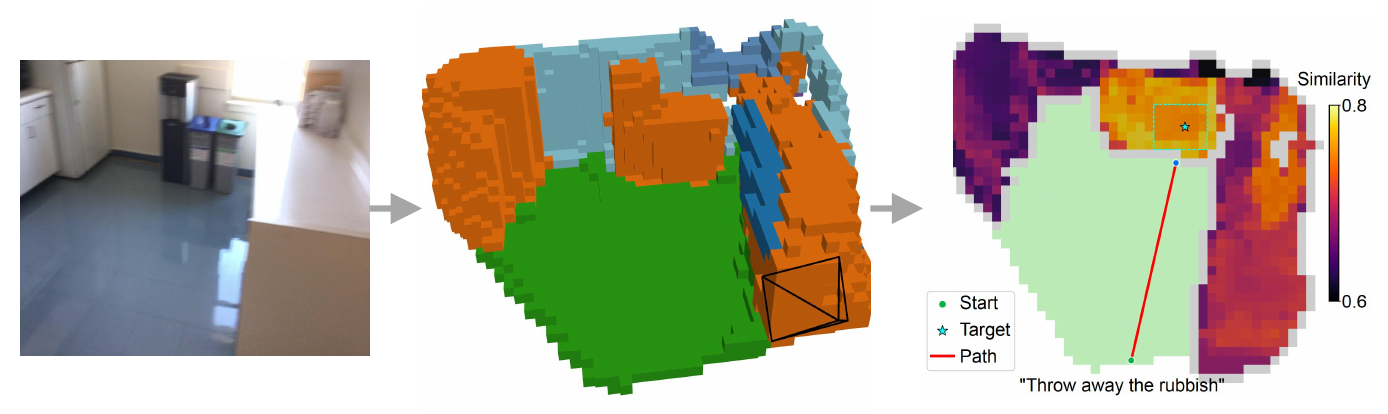}\\[-0.5ex]
\small (b) Language-guided Semantic Navigation
\end{minipage}\\[0.8ex]
\caption{Illustration of open-vocabulary occupancy prediction and its application to language-guided semantic navigation.
(a) Given a monocular image and arbitrary semantic text queries,
the model predicts 3D scene occupancy and assigns semantic labels to occupied voxels.
(b) The resulting semantic occupancy representation supports language-guided navigation, such as planning a path toward targets specified by language instructions, e.g., ''throw away the rubbish''.
}
\label{fig:intro_application}
\vspace{-2mm}
\end{figure}

In addition to geometric complexity, indoor scenes contain diverse, fine-grained, and long-tailed semantic categories.
However, most existing occupancy prediction methods~\cite{monoscene,ISO,gaussianformer} are trained under closed-vocabulary settings with a predefined set of categories.
This restriction limits their ability to recognize rare or previously unseen objects, motivating the need for open-vocabulary occupancy prediction.
Recent advances in Vision-Language Models (VLMs), such as CLIP~\cite{CLIP} and SigLIP~\cite{SigLIP}, have demonstrated remarkable zero-shot transferability by aligning visual representations with natural language.
By transferring language-aligned representations into 3D space, open-vocabulary occupancy prediction enables recognition of arbitrary categories beyond predefined label sets via natural-language queries, as illustrated in Fig.~\ref{fig:intro_application}.

In this work, we study open-vocabulary monocular occupancy prediction for indoor scenes and propose OVM-Occ.
OVM-Occ employs Language-Embedded Gaussians (LE-Gaussians)~\cite{LEGaussian} as the intermediate scene representation,
and constructs an instance-aware vision-language teacher that integrates visual appearance cues with language semantic anchors, 
enabling instance-level feature distillation and open-vocabulary embedding alignment.
The main contributions of this paper can be summarized as follows:
\begin{itemize}
\item We propose OVM-Occ, a Gaussian-based open-vocabulary occupancy framework, 
with a geometry-aware deformable attention mechanism to refine Gaussian semantic features and a geometry-aware decoupled semantic distillation strategy to mitigate interference between geometry learning and semantic learning.
\item We introduce an instance-aware vision-language teacher that provides reliable instance-level semantic supervision via confidence-calibrated vision-language fusion.
\item We introduce OpenOcc-ScanNet and OpenOcc-Replica, two extended benchmarks with more diverse and fine-grained semantic categories for open-vocabulary monocular occupancy evaluation. 
      Extensive experiments demonstrate that our method outperforms previous state-of-the-art approaches on both Occ-ScanNet and the newly introduced benchmarks.
\end{itemize}

\begin{figure*}[t]
\centering
\includegraphics[width=\textwidth]{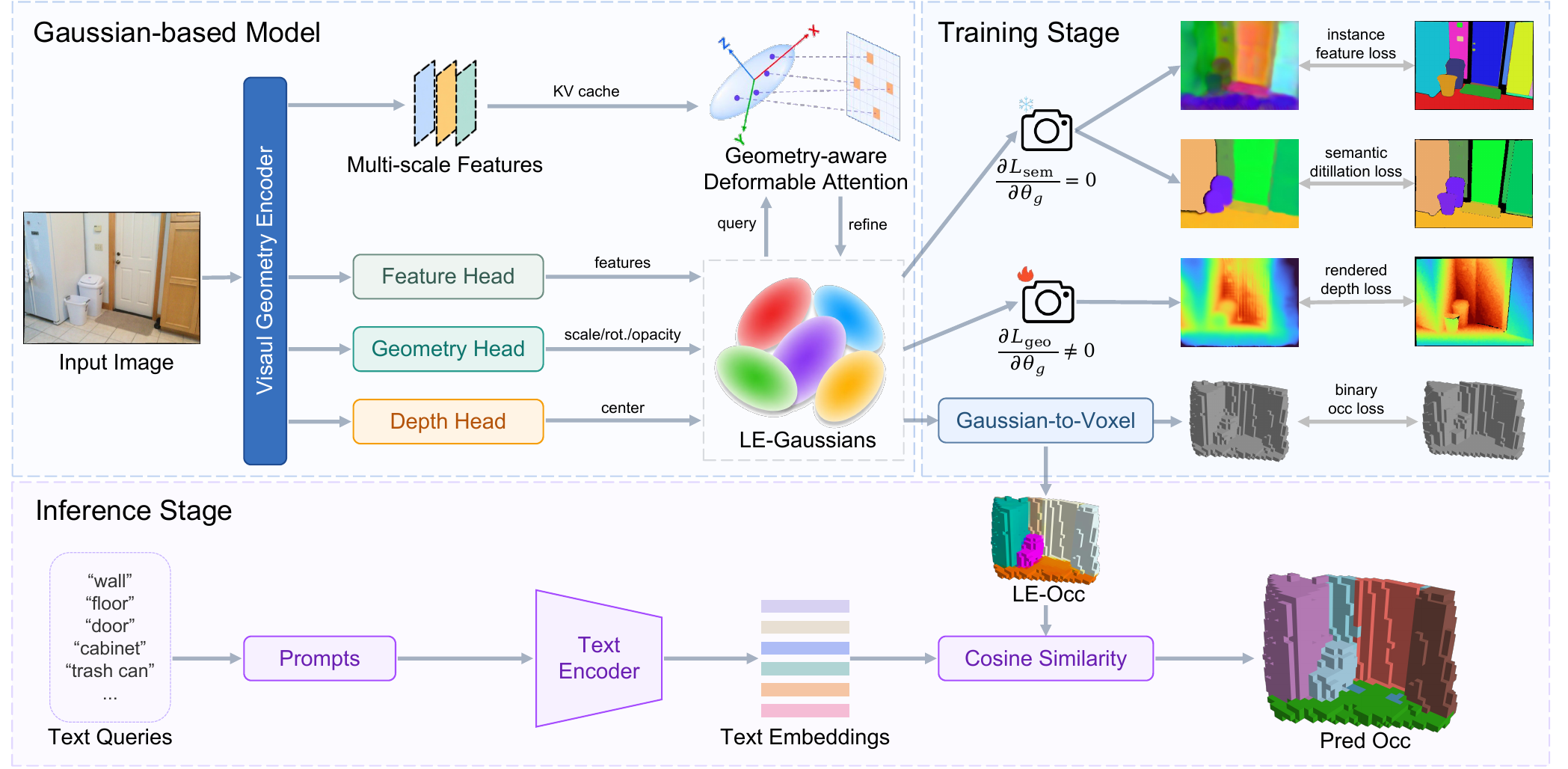}
\caption{Overview of OVM-Occ. 
Given a monocular RGB image, the Gaussian-based model predicts Language-Embedded Gaussians (LE-Gaussians) and refines their features using geometry-aware deformable attention.
During training, the LE-Gaussians are rendered into camera views for 2D semantic and depth supervision, while being aggregated into voxel representations for 3D geometric occupancy supervision.
At inference, the LE-Gaussians are aggregated into Language-Embedded Occupancy (LE-Occ),
where the occupancy features are matched with query text embeddings via cosine similarity to produce open-vocabulary occupancy predictions.
}
\label{fig:pipeline}
\end{figure*}

\section{Related Work}

\subsection{Monocular Occupancy Prediction}
Multi-camera occupancy prediction has been extensively studied in autonomous driving and has achieved remarkable performance~\cite{bevformer,tpvformer,surroundocc,gaussianformer}.
Motivated by embodied perception and advances in visual geometry foundation models~\cite{DepthAnythingv2,VGGT}, recent studies have explored monocular occupancy prediction for indoor scene understanding.
MonoScene~\cite{monoscene} is a pioneering work in monocular 3D occupancy prediction.
ISO~\cite{ISO} leverages monocular depth estimation to lift image features into dense 3D volumes for occupancy prediction. 
Recent works~\cite{embodiedocc,embodiedocc++,GPOcc,roboocc} employ Gaussian primitives as an intermediate scene representation and optimize occupancy prediction via differentiable rendering.
However, these methods primarily focus on improving geometric reconstruction, while semantic modeling remains restricted to predefined label sets, limiting generalization to unseen categories.

\subsection{Open-Vocabulary Occupancy Prediction}
Open-vocabulary occupancy prediction incorporates semantic knowledge from VLMs, enabling semantic understanding beyond predefined categories.
POP-3D~\cite{POP3D} and AGO~\cite{AGO} establish pixel-to-voxel correspondences through geometric projection and distill the corresponding visual embeddings into voxel representations for 3D occupancy prediction.
LOcc~\cite{locc} employs a 2D open-vocabulary semantic segmentation model to generate category-specific masks and distills the corresponding text embeddings into voxel representations.
However, these methods are primarily designed for outdoor driving scenarios with multi-view cameras, and their direct transfer to indoor monocular settings leads to inferior performance~\cite{embodiedocc}.
LegoOcc~\cite{legoocc} extends open-vocabulary occupancy prediction to indoor monocular scenes under geometry-only supervision. 
It introduces 3D LE-Gaussians to jointly model scene geometry and semantics, renders Gaussian semantic features into the image plane, and distills them with vision-language embeddings.
Overall, research on open-vocabulary occupancy prediction for indoor monocular scenes remains relatively underexplored.

\section{Methodology}
This section details the proposed OVM-Occ framework, whose overall architecture is illustrated in Fig.~\ref{fig:pipeline}.

\subsection{Problem Definition}
Given a monocular RGB image $\mathbf I \in \mathbb{R}^{H \times W \times 3}$, the goal is to predict a voxel-wise semantic occupancy field $\mathbf{O} \in \mathbb{R}^{X \times Y \times Z \times N_c}$,
where $X$, $Y$, and $Z$ denote the spatial resolution, and $N_c$ is the number of semantic categories.
During training, only binary occupancy labels and depth observations are used, without voxel-level semantic annotations.
Semantic categories are specified only at inference time through text queries.

\subsection{Language-Embedded Gaussian Representation}
Building on previous works~\cite{Langsplat,LEGaussian,legoocc}, we adopt LE-Gaussians as the intermediate scene representation.
Each LE-Gaussian is parameterized as:
\begin{equation}
\mathcal{G}=\{\boldsymbol{\mu},\mathbf{s},\mathbf{r},\alpha,\mathbf{f}\},
\end{equation}
where $\boldsymbol{\mu}\in\mathbb{R}^3$, $\mathbf{s}\in\mathbb{R}^3$, $\mathbf{r}\in\mathbb{R}^3$, $\alpha\in[0,1]$, and $\mathbf{f}\in\mathbb{R}^d$ denote the Gaussian center, scale, rotation, opacity, and language-aligned feature, respectively.

Following GPOcc~\cite{GPOcc}, we initialize surface Gaussian centers from the predicted depth and extend them along camera rays to construct volumetric Gaussians:
\begin{equation}
\boldsymbol{\mu}_n(p)=\left(\hat{D}(p)+\Delta_n(p)\right)\mathbf d(p),\quad n=0,\ldots,N-1
\end{equation}
where $\mathbf{d}(p)$ denotes the unit camera ray at pixel $p$, $\hat{D}(p)$ and $\Delta_n(p)$ represent the predicted depth and depth offset, and $N$ represents the total number of Gaussians sampled per ray.

We introduce a geometry-aware deformable attention module to refine Gaussian semantic features through geometry-guided feature sampling and aggregation.
We exploit the explicit geometric properties of 3D Gaussians to perform structured sampling.
Specifically, four fixed samples are placed along the two major axes, complemented by two learnable samples that adapt to local visual cues.
The sampling offset $\boldsymbol{\epsilon}_j\in\mathbb{R}^3$ is defined in the local coordinate system, transformed by the Gaussian scale and rotation into a 3D sampling point, and projected onto the 2D image plane:
\begin{equation}
    \mathbf{u}_j = \pi\left( \boldsymbol{\mu} + \mathbf{R}\mathrm{diag}(\mathbf{s})\boldsymbol{\epsilon}_j \right),
\end{equation}
where $\mathbf{u}_j$ denotes the 2D sampling location, $\mathbf{R}$ is the rotation matrix derived from $\mathbf{r}$,
and $\pi(\cdot)$ is the projection function.
For each sample, an attention weight is adaptively predicted by a multi-layer perceptron (MLP) based on the sampled features and geometric cues:
\begin{equation}
  \tilde{w}_j =
  \operatorname{MLP}
  \left(
  \operatorname{Concat}
  \left(
  \mathbf{q}_g,\,
  \mathbf{k}_j,\,
  \cos(\mathbf{q}_g, \mathbf{k}_j),\,
  d_g,\,
  \Delta d_j
  \right)
  \right)
\end{equation}
\begin{equation}
    w_j = \operatorname{Softmax}(\tilde{\mathbf{w}})_j,
\end{equation}
where $\mathbf{q}_g$ denotes the Gaussian query feature, 
$\mathbf{k}_j$ denotes the sampled image features, 
$d_g$ is the Gaussian depth, 
and $\Delta d_j$ is the depth difference between the predicted depth of the sampled pixel and the Gaussian depth.
The sampled features are then aggregated using the attention weights:
\begin{equation}
    \mathbf{h}
    = \sum_{l=1}^{L}\sum_{j=1}^{J}
      w_{l,j}\,
      \mathbf{F}_{l}\!\left(
      \mathbf{u}_{l,j}
      \right),
\end{equation}
where $\mathbf{F}_{l}$ denotes the $l$-th feature map, 
and $\mathbf{h}$ denotes the aggregated feature, which is further projected by an MLP to yield the language-aligned feature $\mathbf f$.

Following LegoOcc~\cite{legoocc}, we adopt the Poisson-based splatting strategy to project LE-Gaussians into voxel-wise occupancy representations with language-aligned features.

\subsection{Instance-Aware Vision-Language Teacher}
\begin{figure}[t]
\centering
\includegraphics[width=0.95\columnwidth]{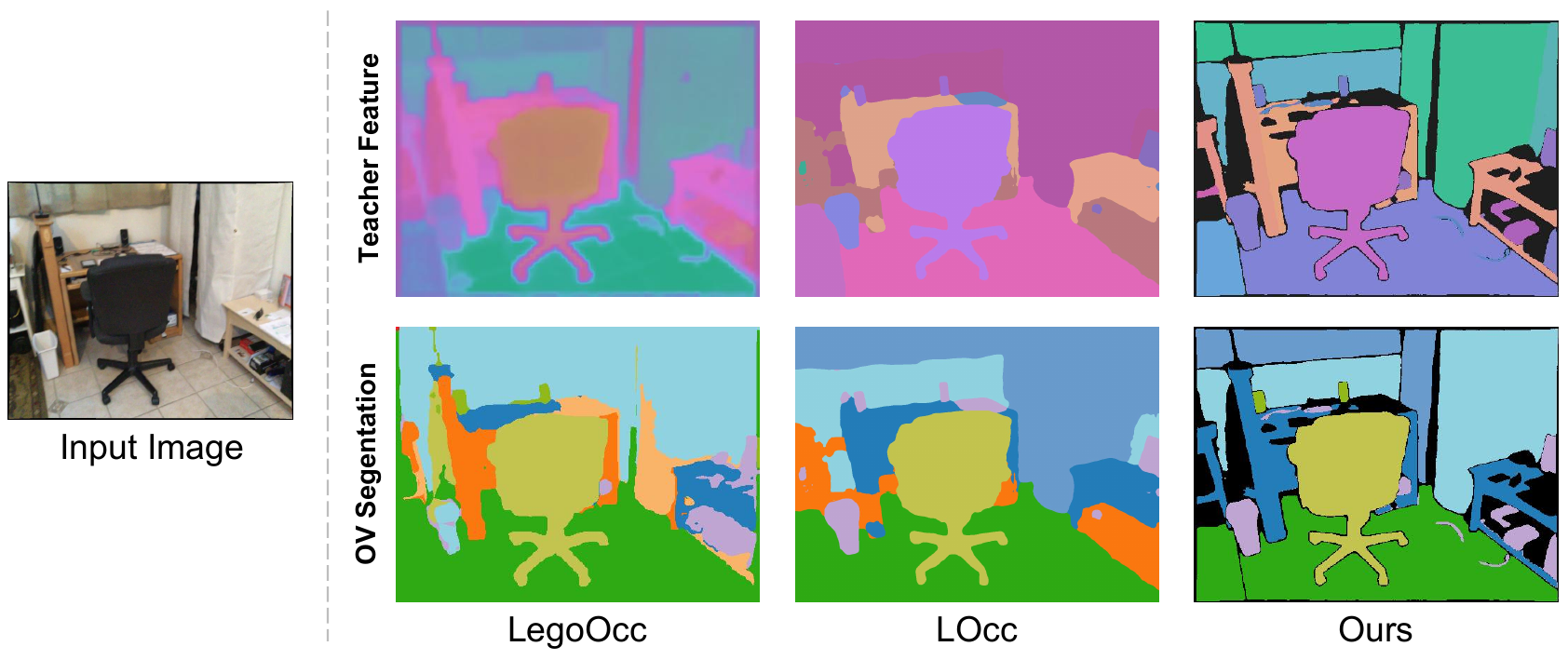}
\caption{Comparison of different teacher embeddings and their corresponding open-vocabulary semantic segmentations.
Top: PCA visualization of teacher embeddings.
Each method is visualized independently; therefore, colors are only comparable within each visualization.
Bottom: Corresponding open-vocabulary segmentation results.
The color legend is provided in Table~\ref{tab:sota_occscannet}.
}
\vspace{-2mm}
\label{fig:TeacherCmp}
\end{figure}

\begin{figure}[t]
\centering
\includegraphics[width=0.95\columnwidth]{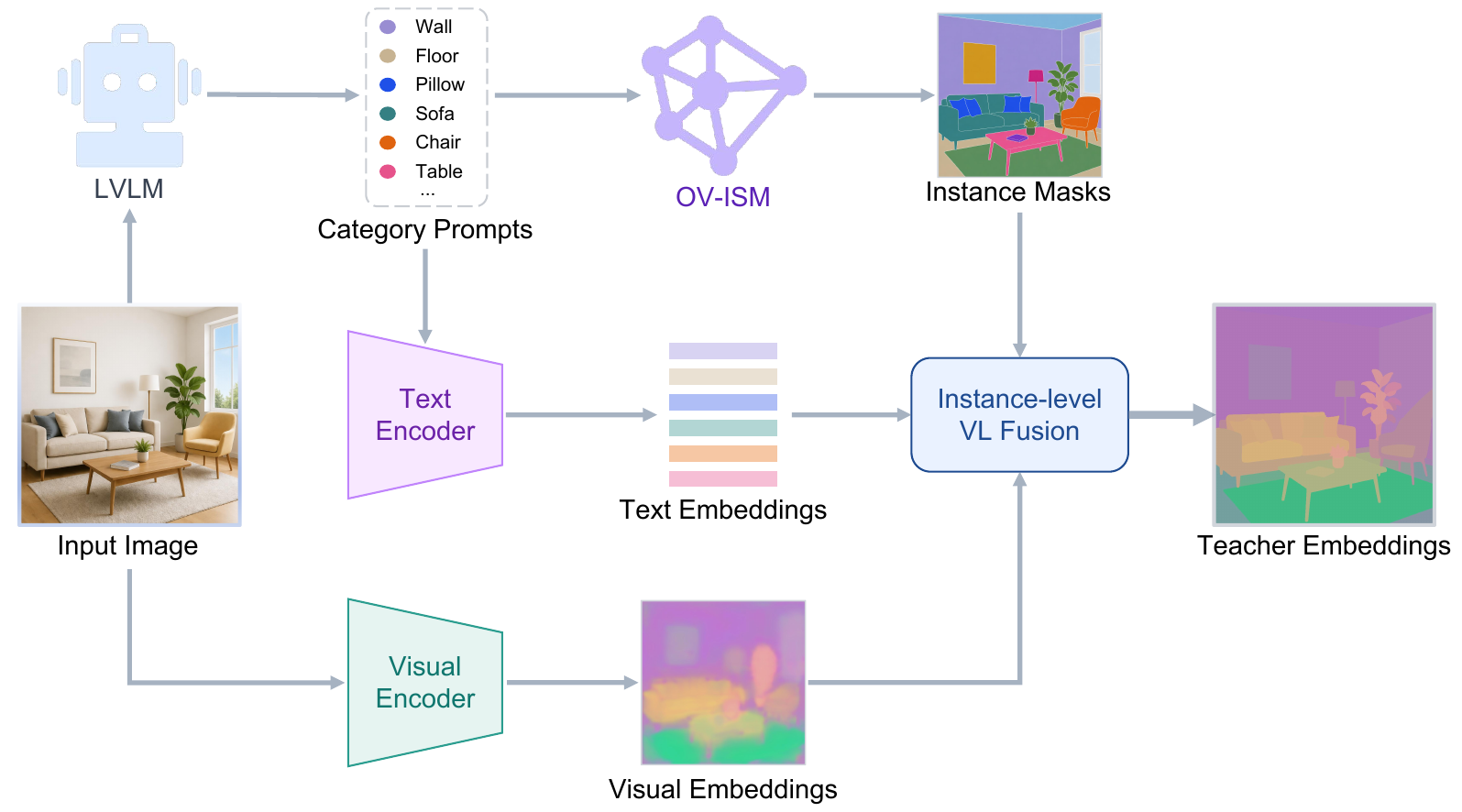}
\caption{Construction of the instance-aware vision-language teacher.
A large vision-language model (LVLM) extracts semantic categories from the input image, 
which are then fed into an open-vocabulary instance segmentation model (OV-ISM) to generate instance masks and into a text encoder to produce text embeddings. 
In parallel, a visual encoder extracts dense visual embeddings from the input image. 
The text and visual embeddings are fused, guided by the instance masks, to obtain instance-aware teacher embeddings.}
\vspace{-2mm}
\label{fig:teacher_construction}
\end{figure}

Existing open-vocabulary occupancy prediction methods~\cite{POP3D,AGO,legoocc} typically rely on dense visual embeddings extracted from CLIP-based models as teacher supervision.
However, these embeddings are learned through image-level contrastive pre-training~\cite{CLIP} and inevitably encode background context and irrelevant regions, thereby yielding noisy semantic supervision.
In contrast, LOcc~\cite{locc} employs language teacher embeddings together with open-vocabulary semantic masks.
While the language embeddings provide relatively clean category-level supervision, 
the semantic masks lack explicit instance awareness, leading to ambiguous supervision around instance boundaries.
Moreover, LOcc ignores rich visual appearance cues, which are crucial for open-vocabulary tasks.
Fig.~\ref{fig:TeacherCmp} provides a qualitative comparison of the aforementioned teachers and their corresponding open-vocabulary segmentation results.
To leverage the complementary strengths of both modalities, we construct an instance-aware vision-language teacher, 
where language embeddings serve as semantic anchors while visual embeddings provide instance-specific appearance cues, 
as shown in Fig.~\ref{fig:teacher_construction}.

We employ a large vision-language model (LVLM) to extract semantic categories from the input image. 
The predicted categories are subsequently refined through text denoising and temporal completion, yielding the final category set $\mathcal{C}=\{c_i\}_{i=1}^{K}$.
The category set is then encoded by the CLIP text encoder to obtain language embeddings $\mathcal{T}=\{\mathbf{t}_i\}_{i=1}^{K}$. 
Meanwhile, the categories are used as text prompts for an open-vocabulary instance segmentation model (OV-ISM) to generate instance masks $\mathcal{M}=\{M_i\}_{i=1}^{M}$.
Consequently, each instance mask $M_i$ is naturally associated with the language embedding of its corresponding text prompt, denoted as $\hat{\mathbf{t}}_i$.

The vision branch employs a CLIP-based visual encoder to extract pixel-wise embeddings $\mathbf{F}$, which are aligned with the text embedding space.
Given the instance masks $\mathcal{M}$, we perform masked average pooling over the dense embeddings $\mathbf{F}$ to obtain instance-level visual embeddings $\mathcal{V}=\{\mathbf{v}_i\}_{i=1}^M$:
\begin{equation}
\mathbf{v}_i=
\frac{\sum_{p}M_i(p)\mathbf{F}(p)}
{\sum_{p}M_i(p)},
\end{equation}
where $M_i(p)\in\{0,1\}$ indicates whether spatial location $p$ belongs to the $i$-th instance mask, and $\mathbf{F}(p)$ denotes the corresponding visual feature.

For each instance mask $M_i$, we apply L2 normalization to the visual embedding $\mathbf{v}_i$ and its corresponding language embedding $\hat{\mathbf{t}}_i$, and then construct a fused representation:
\begin{equation}
    \tilde{\mathbf{e}}_i = \lambda_\text{vis} \mathbf{v}_i + (1-\lambda_\text{vis})\hat{\mathbf{t}}_i ,
\end{equation}
where $\lambda_\text{vis}$ denotes the weighting coefficient of the visual modality.
The cross-modal consistency between the visual and language embeddings is measured by cosine similarity:
\begin{equation}
s_i=\max(\mathbf{v}_i^{\top}\hat{\mathbf{t}}_i, 0).
\end{equation}

The fused representation is normalized to preserve its direction, while the confidence score modulates its magnitude, yielding the final instance-aware vision-language teacher embedding:
\begin{equation}
\mathbf{e}_i=
s_i
\frac{\tilde{\mathbf{e}}_i}
{\|\tilde{\mathbf{e}}_i\|_2}.
\end{equation}

\subsection{Geometry-Aware Decoupled Semantic Distillation}
\noindent\textbf{Semantic-to-geometry decoupling.}
LegoOcc~\cite{legoocc} renders Gaussian semantic features onto the image plane and applies feature supervision.
During optimization, the gradients from the feature rendering loss are propagated to the geometry branch.
As a result, the optimization may tend to reduce the rendering loss by adjusting Gaussian geometry rather than improving semantic representations.
Meanwhile, the high-dimensional CLIP teacher embeddings inevitably contain noisy semantic responses, introducing unreliable supervision signals into the geometry branch.

To address this issue, we detach the gradients of Gaussian geometry parameters in feature rendering optimization:
\begin{equation}
\frac{\partial \mathcal{L}_{\mathrm{sem}}}{\partial \theta_g}=0
\end{equation}
where $\theta_g=\{\boldsymbol{\mu},\mathbf{s},\mathbf{r},\alpha\}$ denotes the geometry parameters.

\noindent\textbf{Geometry-to-semantic decoupling.}
Inaccurate depth estimation may cause misalignment between projected Gaussian features and image features, resulting in unreliable semantic supervision.
We apply a depth-error-aware mask to filter geometrically unreliable pixels during semantic distillation:
\begin{equation}
    M^{d} = \mathbb{I}\left(|\hat{D} - {D}^{\mathrm{gt}}| < \tau_d\right),
\end{equation}
where $\hat{D}$ and $D^{\mathrm{gt}}$ denote the predicted and ground-truth depth maps, respectively,
and $\tau_d$ is the depth error threshold.

\subsection{Training Objective}
\noindent\textbf{Geometry supervision.}
We employ the binary focal loss $\mathcal{L}_\text{focal}$ and the Lovász loss $\mathcal{L}_\text{lov}$~\cite{lovasz} to supervise voxel-wise 3D occupancy prediction.
The Huber loss $\mathcal{L}_{\text{depth}}$ is adopted for dense depth estimation in the input view and depth rendering in neighboring views.




\noindent\textbf{Semantic supervision.}
We render Gaussian features into the 2D image plane of the input and neighboring views for supervision.
The rendered semantic features are optimized using a confidence-weighted distillation loss $\mathcal{L}_{\text{sem}}$ based on cosine similarity with the teacher embeddings:
\begin{equation}
\mathcal{L}_{\mathrm{sem}}
=
\frac{
\sum_{i=1}^{M}\sum_{p}
s_i M_i(p) M^d(p)
\left[
1-\cos\left(\mathbf f(p),\mathbf e_i\right)
\right]
}{
\sum_{i=1}^{M}\sum_{p}s_i M_i(p) M^d(p)
}.
\end{equation}
Additionally, an auxiliary instance feature contrastive loss $\mathcal{L}_{\text{ins}}$ is introduced to enhance instance-level consistency.


The overall training objective is formulated as:
\begin{equation}
\mathcal{L} =
\lambda_{1}\mathcal{L}_{\text{focal}}
+\lambda_{2}\mathcal{L}_{\text{lov}}
+\lambda_{3}\mathcal{L}_{\text{depth}}
+\lambda_{4}\mathcal{L}_{\text{sem}}
+\lambda_{5}\mathcal{L}_{\text{ins}},
\end{equation}
where $\lambda_i$ denotes the balancing weight for each loss term.

\section{Experiments}

\begin{table*}[t]
\centering
\caption{Comparison of monocular occupancy prediction performance on the Occ-ScanNet benchmark. 
Closed-vocabulary methods are trained with both semantic and occupancy ground-truth annotations, 
whereas open-vocabulary methods use only occupancy ground truth.
The best result among open-vocabulary methods is highlighted in bold.}
\label{tab:sota_occscannet}
\footnotesize
\setlength{\tabcolsep}{1.5pt}
\begin{tabular}{l|C|*{11}{C}|C}
\toprule
\raisebox{-2em}{Method} & \headercell{0.82cm}{IoU} & \vclassheader{ceilingcolor}{Ceiling} & \vclassheader{floorcolor}{Floor} & \vclassheader{wallcolor}{Wall} & \vclassheader{windowcolor}{Window} & \vclassheader{chaircolor}{Chair} & \vclassheader{bedcolor}{Bed} & \vclassheader{sofacolor}{Sofa} & \vclassheader{tablecolor}{Table} & \vclassheader{tvscolor}{TVs} & \vclassheader{furniturecolor}{Furniture} & \vclassheader{objectscolor}{Objects} & \headercell{0.82cm}{mIoU} \\
\midrule
\multicolumn{14}{l}{\textit{Closed-vocabulary: supervised by full annotations.}} \\
\midrule    
TPVFormer~\cite{tpvformer} & 33.39 & 6.96 & 32.97 & 14.41 & 9.10 & 24.01 & 41.49 & 45.44 & 28.61 & 10.66 & 35.37 & 25.31 & 24.94 \\
GaussianFormer~\cite{gaussianformer} & 40.91 & 20.70 & 42.00 & 23.40 & 17.40 & 27.00 & 44.30 & 44.80 & 32.70 & 15.30 & 36.70 & 25.00 & 29.93 \\
MonoScene~\cite{monoscene} & 41.60 & 15.17 & 44.71 & 22.41 & 12.55 & 26.11 & 27.03 & 35.91 & 28.32 & 6.57 & 32.16 & 19.84 & 24.62 \\
ISO~\cite{ISO} & 42.16 & 19.88 & 41.88 & 22.37 & 16.98 & 29.09 & 42.43 & 42.00 & 29.60 & 10.62 & 36.36 & 24.61 & 28.71 \\
SurroundOcc~\cite{surroundocc} & 42.52 & 18.90 & 49.30 & 24.80 & 18.00 & 26.80 & 42.00 & 44.10 & 32.90 & 18.60 & 36.80 & 26.90 & 30.83 \\
EmbodiedOcc~\cite{embodiedocc} & 53.55 & 39.60 & 50.40 & 41.40 & 31.70 & 40.90 & 55.00 & 61.40 & 44.00 & 36.10 & 53.90 & 42.20 & 45.15 \\
EmbodiedOcc++~\cite{embodiedocc++} & 54.90 & 36.40 & 53.10 & 41.80 & 34.40 & 42.90 & 57.30 & 64.10 & 45.20 & 34.80 & 54.20 & 44.10 & 46.20 \\
RoboOcc~\cite{roboocc} & 56.48 & 45.36 & 53.49 & 44.35 & 34.81 & 43.38 & 56.93 & 63.35 & 46.35 & 36.12 & 55.48 & 44.78 & 47.67 \\
GPOcc-DPT~\cite{GPOcc} & 56.96 & 51.42 & 50.35 & 46.97 & 41.84 & 46.98 & 60.39 & 66.16 & 50.51 & 47.97 & 58.88 & 49.23 & 51.88 \\
\midrule
\rowcolor{gray!12}
\multicolumn{14}{l}{\textit{Open-vocabulary: supervised by geometry-only annotations.}} \\
\midrule
\rowcolor{gray!12}
POP-3D~\cite{POP3D} & 35.32 & 0.00 & 23.89 & 16.59 & 0.01 & 4.62 & 8.65 & 2.01 & 7.61 & 0.00 & 2.14 & 0.08 & 5.96 \\
\rowcolor{gray!12}
LOcc~\cite{locc} & 36.70 & 0.00 & 24.40 & 17.37 & 1.83 & 9.31 & 10.73 & 23.42 & 11.29 & 0.01 & 3.15 & 0.19 & 9.25 \\
\rowcolor{gray!12}
LegoOcc~\cite{legoocc} & 59.50 & 17.81 & 41.65 & 28.52 & 18.91 & 19.32 & 33.05 & 34.92 & 19.16 & \textbf{5.36} & 5.88 & 6.94 & 21.05 \\
\rowcolor{gray!12}
\textbf{OVM-Occ (Ours)} & \textbf{60.10} & \textbf{44.82} & \textbf{52.76} & \textbf{36.78} & \textbf{19.38} & \textbf{33.22} & \textbf{52.07} & \textbf{50.90} & \textbf{30.92} & 3.88 & \textbf{25.63} & \textbf{15.33} & \textbf{33.25} \\
\bottomrule
\end{tabular}
\end{table*}

\subsection{Experimental Setup}
\noindent\textbf{Datasets.}
We evaluate our method on the ScanNet~\cite{scannet} and Replica~\cite{Replica} datasets.
The Occ-ScanNet~\cite{ISO} benchmark provides annotations for only 11 semantic classes, limiting its ability to support open-vocabulary occupancy evaluation.
To address this issue, we construct the OpenOcc-ScanNet and OpenOcc-Replica benchmarks with fine-grained semantic annotations,
yielding four evaluation protocols: OpenOcc-ScanNet11, OpenOcc-ScanNet20, OpenOcc-ScanNet40, and OpenOcc-Replica35.

\noindent\textbf{Evaluation metrics.}
Following prior work~\cite{ISO,embodiedocc,legoocc}, we use overall occupancy IoU and mean IoU (mIoU) over all semantic categories as the evaluation metrics.

\noindent\textbf{Implementation details.}
For teacher construction, we use Qwen2.5-VL-3B~\cite{qwen} for semantic category extraction, 
CLIP text encoder for text embedding extraction,
SAM 3~\cite{sam3} for open-vocabulary instance segmentation, 
and OpenSeg~\cite{openseg} for visual embedding extraction.
For the student model, following~\cite{embodiedocc,GPOcc,legoocc}, we employ a fine-tuned DepthAnything-V2~\cite{DepthAnythingv2} as the backbone.
The network is optimized using AdamW with a weight decay of 0.01.
The initial learning rate is set to $2\times10^{-4}$ and follows a cosine annealing schedule.
The model is trained for 10 epochs with a batch size of 8 on 4 NVIDIA RTX 5090 GPUs.
For key hyperparameters, 
we set the number of Gaussians sampled per ray to $N=8$, 
the visual weight $\lambda_\text{vis}=0.3$,
the depth error threshold $\tau_d=0.3$,
and the loss weights $\lambda_1=100.0,\lambda_2=1.0,\lambda_3=2.0,\lambda_4=2.0,\lambda_5=0.5$.
More implementation details will be provided in the released code.

\subsection{Main Results}


\begin{figure*}[t]
\centering
\includegraphics[width=0.91\textwidth,trim=0 14bp 0 0,clip]{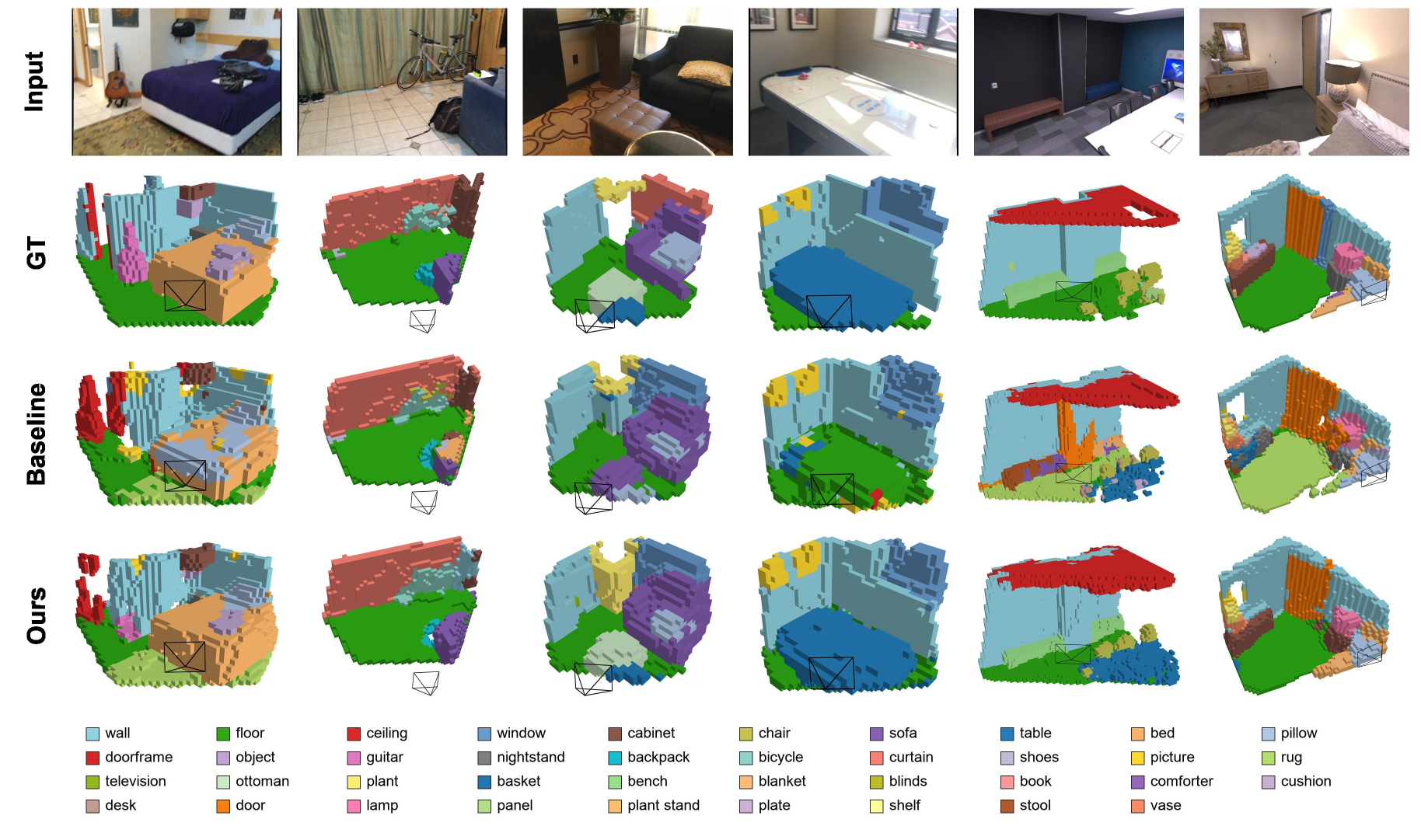}
\captionsetup{skip=2pt}
\caption{Qualitative results on the OpenOcc-ScanNet and OpenOcc-Replica benchmarks with more diverse semantic categories.}
\label{fig:compare_openocc}
\vspace{-1mm}
\end{figure*}

\begin{table*}[t]
\centering
\caption{Comparison of monocular occupancy prediction performance on the OpenOcc-ScanNet benchmark. 
Per-class IoU is reported for a subset of semantic categories under the OpenOcc-ScanNet40 protocol.}
\label{tab:openocc-scannet}
\footnotesize
\setlength{\tabcolsep}{1pt}
\begin{tabular}{l@{\hspace{2pt}}|@{\hspace{2pt}}C@{\hspace{2pt}}|@{\hspace{2pt}}*{10}{D}@{\hspace{3pt}}|@{\hspace{2pt}}c@{\hspace{2pt}}c@{\hspace{2pt}}c}
\toprule
& & \multicolumn{10}{c@{\hspace{-2pt}\vrule\hspace{2pt}}}{per-class IoU} & \multicolumn{3}{c}{mIoU} \\[0.6ex]
Method & IoU & Cabinet & Counter & Pillow & Books & Towel & Mirror & Fridge & Toilet & Sink & Bathtub & 11-class & 20-class & 40-class \\
\midrule
POP-3D~\cite{POP3D} & 48.94 & 0.28 & 4.54 & 0.11 & 0.00 & 9.28 & 0.00 & 7.97 & 9.79 & 11.02 & 1.07 & 13.22 & 11.37 & 6.19 \\
LOcc~\cite{locc} & 44.70 & 17.30 & \textbf{26.17} & 0.00 & 0.00 & 0.00 & 0.00 & 17.39 & 14.25 & 9.84 & 37.87 & 18.89 & 21.86 & 12.91 \\
LegoOcc~\cite{legoocc} & 56.22 & 8.73 & 9.10 & 9.13 & 10.63 & 6.15 & 0.00 & 4.96 & 13.82 & 17.13 & 33.41 & 23.77 & 20.64 & 12.93 \\
\textbf{OVM-Occ (Ours)} & \textbf{56.82} & \textbf{22.77} & 21.44 & \textbf{28.82} & \textbf{19.68} & \textbf{18.62} & \textbf{24.06} & \textbf{19.56} & \textbf{32.06} & \textbf{38.18} & \textbf{57.38} & \textbf{32.40} & \textbf{30.83} & \textbf{21.79} \\
\bottomrule
\end{tabular}
\vspace{-2mm}
\end{table*}

\begin{table}[t]
\centering
\caption{Open-vocabulary occupancy prediction results on OpenOcc-Replica35 under different training settings.
Each entry reports IoU / mIoU.}
\label{tab:pretrain_data_scale}
\footnotesize
\setlength{\tabcolsep}{5pt}
\begin{tabular}{lccc}
\toprule
Method & \shortstack{From Scratch\\1.1K frames} & \shortstack{Pre-trained\\100 frames} & \shortstack{Pre-trained\\1.1K frames} \\
\midrule
LegoOcc & 62.85 / 9.27 & 47.76 / 8.12 & 73.65 / 14.90 \\
\textbf{Ours} & \textbf{64.21 / 14.62} & \textbf{50.09 / 14.44} & \textbf{77.88 / 23.28} \\
\bottomrule
\end{tabular}
\vspace{-2mm}
\end{table}

We compare OVM-Occ with representative open-vocabulary methods, including POP-3D~\cite{POP3D}, LOcc~\cite{locc}, and LegoOcc~\cite{legoocc}.
Table~\ref{tab:sota_occscannet} reports the quantitative comparison results on the Occ-ScanNet benchmark.
OVM-Occ achieves state-of-the-art performance among all open-vocabulary methods, 
with 60.10 IoU and 33.25 mIoU, outperforming LegoOcc by 0.60 and 12.20, respectively.
We further evaluate the open-vocabulary generalization ability on the OpenOcc-ScanNet benchmark, which contains more diverse and fine-grained semantic categories.
As shown in Table~\ref{tab:openocc-scannet}, our method consistently outperforms baselines under all evaluation protocols.
Notably, OVM-Occ maintains substantial performance gains even under the challenging 40-class protocol, demonstrating the robustness of the proposed method under a significantly expanded semantic label space.

We evaluate few-shot transfer performance on the OpenOcc-Replica benchmark.
As shown in Table~\ref{tab:pretrain_data_scale}, our method consistently outperforms LegoOcc across all training settings.
With pre-training on Occ-ScanNet, our method achieves comparable semantic performance using only 100 frames from Replica, reaching 14.44 mIoU versus 14.62 mIoU when trained from scratch on 1.1K frames.
With 1.1K frames for fine-tuning, our method further improves to 77.88 IoU and 23.28 mIoU.

 
Fig.~\ref{fig:compare_openocc} presents qualitative comparisons on the OpenOcc-ScanNet and OpenOcc-Replica benchmarks with the expanded semantic label space.
Compared with LegoOcc, 
OVM-Occ achieves more accurate semantic predictions with fewer category confusions and stronger instance-level consistency, 
reducing fragmented predictions within individual objects.
These improvements highlight the effectiveness of the proposed instance-aware vision-language supervision.
Moreover, OVM-Occ successfully recognizes long-tail categories (e.g., guitar and bicycle), demonstrating its open-vocabulary generalization capability beyond frequent categories.

\subsection{Ablation Study}
\paragraph{Factorized ablation of teacher embeddings and student frameworks}
Table~\ref{tab:teacher_student} presents a factorized ablation by combining different teacher embeddings with different student frameworks.
Compared with the Trident~\cite{trident} teacher, 
the proposed instance-aware vision-language teacher consistently improves semantic prediction by over 8 mIoU points for both LegoOcc and our framework.
Under the same teacher, our framework consistently surpasses LegoOcc, achieving gains of 3.49 and 3.75 mIoU with the Trident and proposed teacher, respectively.
These results verify the effectiveness of both the proposed teacher and student framework.

\begin{table}[t]
\centering
\caption{Ablation of different teacher embeddings and student frameworks. VL denotes vision-language.}
\label{tab:teacher_student}
\footnotesize
\setlength{\tabcolsep}{0pt}
\begin{tabular*}{0.98\columnwidth}{@{\hspace{2pt}\extracolsep{\fill}}llcc@{\hspace{2pt}}}
\toprule
Student & Teacher & IoU & mIoU \\
\midrule
LegoOcc & Trident Visual Embedding~\cite{trident} & 59.50 & 21.05 \\
LegoOcc & Instance-aware VL Embedding & 59.71 & 29.50 \\
Ours & Trident Visual Embedding~\cite{trident} & 60.07 & 24.54 \\
Ours & Instance-aware VL Embedding & \textbf{60.10} & \textbf{33.25} \\
\bottomrule
\end{tabular*}
\vspace{-3mm}
\end{table}

\paragraph{Ablation of teacher construction}
Table~\ref{tab:ablation_teacher} compares different teacher construction strategies.
The instance-aware vision-language teacher achieves the best performance, 
outperforming the pixel-level vision and category-level language teachers by 5.09 and 1.97 mIoU points, respectively.
These results demonstrate that the proposed teacher construction effectively exploits the complementary strengths of visual appearance cues and language semantic anchors.

\begin{table}[t]
\centering
\caption{Ablation of teacher construction.}
\label{tab:ablation_teacher}
\small
\setlength{\tabcolsep}{5pt}
\begin{tabular}{>{\raggedright\arraybackslash}p{0.5\columnwidth}cc}
\toprule
Teacher Construction & IoU & mIoU \\
\midrule
Pixel-level Vision  & \textbf{60.20} & 28.16 \\
Category-level Language  & 60.09 & 31.28 \\
Instance-aware Vision-Language & 60.10 & \textbf{33.25} \\
\bottomrule
\end{tabular}
\end{table}


\paragraph{Ablation of geometry-aware feature refinement}

\begin{table}[t]
\centering
\caption{Ablation of feature refinement modules.}
\label{tab:geometry_attention}
\small
\setlength{\tabcolsep}{3pt}
\begin{tabular}{>{\raggedright\arraybackslash}p{0.6\columnwidth}cc}
\toprule
Refinement Module & IoU & mIoU \\
\midrule
None & 59.57 & 31.98 \\
Deformable Attention & 60.02 & 32.43 \\
Geometry-aware Deformable Attention & \textbf{60.10} & \textbf{33.25} \\
\bottomrule
\end{tabular}
\vspace{-3mm}
\end{table}

Table~\ref{tab:geometry_attention} reports the ablation results of different feature refinement modules.
In the "None" setting, only an MLP is used to project Gaussian features into the semantic embedding space, achieving 59.57 IoU and 31.98 mIoU.
Introducing a deformable attention module to aggregate information from multi-scale image features leads to improved performance.
The proposed geometry-aware deformable attention further improves to 60.10 IoU and 33.25 mIoU, 
demonstrating that incorporating Gaussian geometric priors enables more accurate feature sampling for Gaussian feature refinement.

\paragraph{Ablation of geometry-aware decoupled semantic distillation}
We evaluate the effectiveness of different geometry-semantic decoupling strategies during feature distillation, as summarized in Table~\ref{tab:geometry_semantic_decoupling}.
Applying only Spatial SG improves mIoU from 31.11 to 32.12, 
indicating that decoupling Gaussian geometry from semantic supervision enables more effective feature distillation.
In contrast, Opacity SG primarily benefits geometric occupancy prediction, increasing IoU from 58.38 to 59.39.
This suggests that decoupling Gaussian opacity prevents the optimization from reducing the distillation loss by suppressing Gaussian opacity, thereby avoiding interference with geometric occupancy optimization.
Combining both stop-gradient strategies achieves further improvements in geometric occupancy and semantic performance, demonstrating their complementary effects in mitigating geometry-semantic interference. 
Finally, incorporating depth-error-aware masks yields additional gains in IoU and mIoU, indicating that filtering out geometrically unreliable regions provides cleaner semantic supervision.

\begin{table}[t]
\centering
\caption{Ablation of geometry-semantic decoupling strategies. Spatial SG and Opacity SG denote stop-gradient operations on Gaussian geometry attributes (mean, rotation, and scale) and opacity, respectively.}
\label{tab:geometry_semantic_decoupling}
\small
\setlength{\tabcolsep}{3pt}
\begin{tabular}{cc>{\centering\arraybackslash}p{0.24\columnwidth}|cc}
\toprule
Spatial SG & Opacity SG & Depth Mask & IoU & mIoU \\
\midrule
 &  &  & 58.38 & 31.11 \\
\cmark &  &  & 58.30 & 32.12 \\
 & \cmark &  & 59.39 & 31.55 \\
\cmark & \cmark &  & 59.86 & 32.51 \\
\cmark & \cmark & \cmark & \textbf{60.10} & \textbf{33.25} \\
\bottomrule
\end{tabular}
\end{table}

\paragraph{Runtime analysis}
\begin{table}[t]
\centering
\caption{Runtime and memory analysis under different numbers of sampled Gaussians $N$ per ray.}
\label{tab:runtime_analysis}
\small
\setlength{\tabcolsep}{5pt}
\begin{tabular}{c|ccc|cc}
\toprule
$N$ & \#Gauss. & Mem. (GB) & Time (ms) & IoU & mIoU \\
\midrule
1  & 4,144  & 1.59 & 32.93  & 58.99 & 31.26 \\
2  & 8,288  & 1.65 & 47.50  & 59.77 & 32.11 \\
4  & 16,576 & 2.22 & 65.12  & 59.99 & 32.59 \\
8  & 33,152 & 4.00 & 81.26 & \textbf{60.10} & \textbf{33.25} \\
16 & 66,304 & 7.56 & 108.84 & 59.76 & 32.84 \\
\bottomrule
\end{tabular}
\vspace{-3mm}
\end{table}

Table~\ref{tab:runtime_analysis} summarizes the runtime and memory consumption under different numbers of sampled Gaussians $N$ along each ray. 
All measurements are conducted on an NVIDIA RTX 5090 GPU.
Increasing $N$ from 1 to 8 consistently improves both IoU and mIoU, 
indicating that denser Gaussian sampling provides a more complete volumetric representation of the scene.
However, the improvement saturates at $N=8$, 
and further increasing $N$ to 16 may introduce redundant samples, leading to slightly degraded performance while substantially increasing runtime and memory consumption.
Therefore, $N=8$ is adopted as the default setting, achieving the optimal trade-off between accuracy and efficiency.
In addition, we deploy OVM-Occ on the NVIDIA Jetson Thor platform, achieving an average inference latency of 141.4 ms per frame.

\subsection{Real-World Experiments}
\begin{figure}[t]
\centering
\includegraphics[width=\columnwidth]{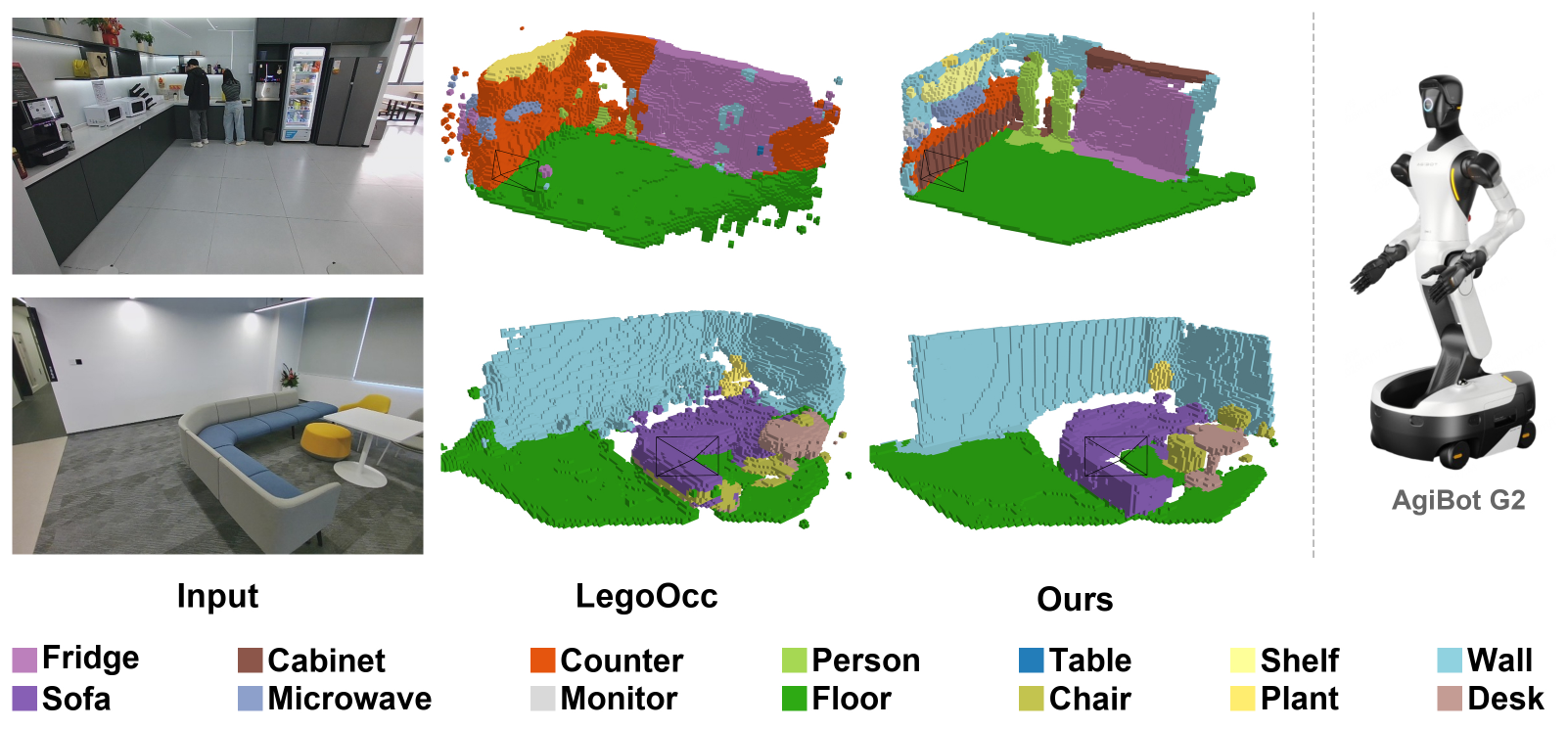}
\caption{Qualitative comparisons between LegoOcc and our method on our real-world dataset collected using an AgiBot G2 robot.}
\label{fig:real_world_comparison}
\vspace{-1mm}
\end{figure}

To evaluate the adaptability of OVM-Occ to real-world environments, 
we collect a monocular RGB dataset comprising 3,260 frames across 5 indoor scenes using a physical robot (AgiBot G2).
We then perform fine-tuning of the Occ-ScanNet-pretrained model using real-world data.
As shown in Table~\ref{tab:real_world}, our method outperforms LegoOcc in both geometric occupancy prediction and semantic segmentation, achieving improvements of 25.47 IoU and 15.76 mIoU, respectively.
In particular, our method achieves better recognition of small objects, such as bags, boxes, and monitors.
The qualitative comparisons in Fig.~\ref{fig:real_world_comparison} further demonstrate the superior performance of our method.

\begin{table}[t]
\centering
\caption{Comparison of monocular occupancy prediction performance on our real-world dataset.}
\label{tab:real_world}
\scriptsize
\setlength{\tabcolsep}{1pt}
\renewcommand{\arraystretch}{1.12}
\begin{tabular}{@{}>{\raggedright\arraybackslash}m{0.14\columnwidth}@{\hspace{2pt}}|@{\hspace{2pt}}c@{\hspace{2pt}}|@{\hspace{2pt}}*{6}{>{\centering\arraybackslash}m{0.095\columnwidth}}@{\hspace{2pt}}|@{\hspace{2pt}}c@{}}
\toprule
 & & \multicolumn{6}{c@{\hspace{2pt}}|@{\hspace{2pt}}}{per-class IoU} & mIoU \\[0.6ex]
Method & IoU & Bag & Box & Cabinet & Monitor & Micro-wave & Trash Can & 20-class \\
\midrule
LegoOcc & 22.14 & 0.72 &  4.26 & 1.70 & 0.00 &  6.61 & 8.69 & 5.50 \\
\textbf{Ours} & \textbf{47.61} & \textbf{12.09} & \textbf{12.71} & \textbf{29.27} & \textbf{19.83} & \textbf{31.01} & \textbf{20.88} & \textbf{21.26} \\
\bottomrule
\end{tabular}
\vspace{-2mm}
\end{table}

\begin{figure}[t]
\centering
\includegraphics[width=\columnwidth]{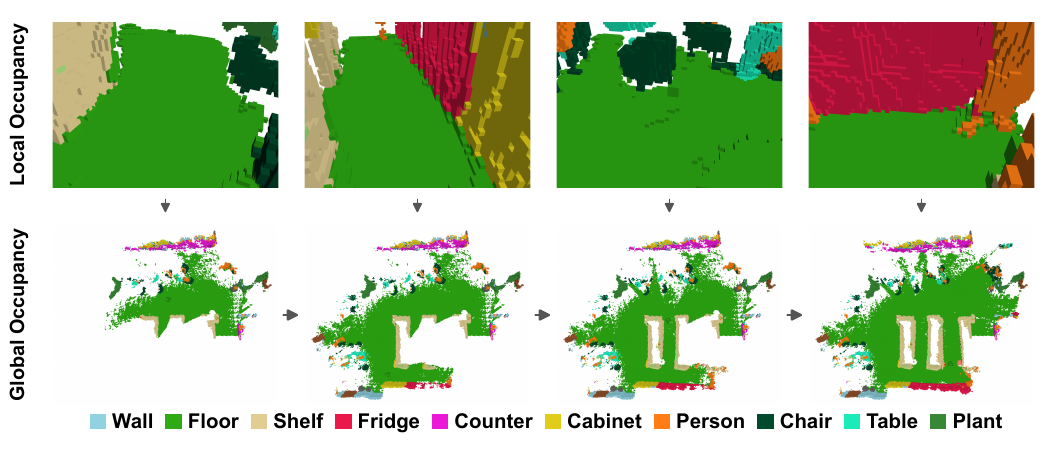}
\caption{Application to multi-frame semantic mapping in a supermarket scene. 
Consecutive local semantic occupancy predictions are accumulated through pose-aligned fusion to build a global 3D semantic occupancy map.}
\label{fig:global_mapping}
\vspace{-2mm}
\end{figure}

To further demonstrate the applicability of OVM-Occ to real-world robotic scenarios, 
we integrate consecutive local occupancy predictions into a global coordinate system through pose alignment and cumulative voting. 
As illustrated in Fig.~\ref{fig:global_mapping}, this process progressively constructs a global 3D semantic occupancy map from sequential observations.

\section{Conclusion}
In this work, we present OVM-Occ, a Gaussian-based framework for open-vocabulary monocular occupancy prediction in indoor scenes.
OVM-Occ introduces an instance-aware vision-language teacher that provides instance-level semantic supervision by integrating visual appearance cues with language semantic anchors.
Furthermore, geometry-aware deformable attention and geometry-aware decoupled semantic distillation enable more effective semantic representation learning.
We establish OpenOcc-ScanNet and OpenOcc-Replica, two benchmarks with more diverse and fine-grained semantic categories.
Extensive experiments demonstrate that our method achieves state-of-the-art performance for open-vocabulary monocular occupancy prediction.

\bibliographystyle{ieeetr}
\bibliography{ovmocc}

\end{document}